\documentclass[a4paper]{article}

\usepackage{graphicx}
\usepackage{enumitem}
\setlist{nolistsep}

\usepackage{multirow} 
\usepackage{graphicx} % more modern
\usepackage{subfigure} 

\usepackage{natbib}

\usepackage{algorithm}
\usepackage{algorithmic}

\usepackage{amsmath}
\usepackage{amsfonts}

\usepackage{hyperref}

\title{Learning invariant features through local space contraction}
       
\author{Salah Rifai, Xavier Muller, Xavier Glorot, Gr\'egoire Mesnil \\ Yoshua Bengio and Pascal Vincent}

\usepackage{url}
\newcommand{\trace}{\mathop{\mathrm{Tr}}}

\newcommand{\R}{\sf{I\!R}}

\newcommand{\E}{\mathbb{E}}
\newcommand{\J}{\mathcal{J}}

\newcommand{\mypm}[1]{{\scriptsize $\pm{#1}$}}
\newcommand{\mnistfull}{{\it MNIST}}
\newcommand{\mnistrotbackimage}{{\it bg-img-rot}}
\newcommand{\mnistbasic}{{\it basic}}
\newcommand{\mnistrot}{{\it rot}}
\newcommand{\mnistbackrand}{{\it bg-rand}}
\newcommand{\mnistbackimage}{{\it bg-img}}
\newcommand{\rectangles}{{\it rect}}
\newcommand{\rectanglesimage}{{\it rect-img}}

\newcommand{\myparagraph}[1]{{\bf #1}}

\newenvironment{myitemize}{
\begin{itemize}
  \setlength{\itemsep}{1pt}
  \setlength{\parskip}{0pt}
  \setlength{\parsep}{0pt}
}{\end{itemize}}

\begin{document}
\maketitle
\thanks{Dept. IRO, Universit\'e de Montr\'eal. Montr\'eal (QC), H3C 3J7, Canada}
\begin{center}          Technical report 1360 \end{center}
\begin{abstract}
% ABSTRACT NEEDS TO BE REWORKED, mainly we our contribution can be divided in two big parts:
% - Analysis of pretraining from a geometrical contraction point of view,
% - CAE inspired from the former analysis beating state of art.

We present in this paper a novel approach for training deterministic
auto-encoders. We show that by adding a well chosen penalty term to the
classical reconstruction cost function, we can achieve results that equal
or surpass those attained by other regularized auto-encoders as well as
denoising auto-encoders on a range of datasets. This penalty term
corresponds to the Frobenius norm of the Jacobian matrix of the encoder
activations with respect to the input.  We show that this penalty term
results in a localized space contraction which in turn yields robust
features on the activation layer.  Furthermore, we show how this penalty
term is related to both regularized auto-encoders and denoising encoders
and how it can be seen as a link between deterministic and
non-deterministic auto-encoders. We find empirically that this penalty
helps to carve a representation that better captures the local directions
of variation dictated by the data, corresponding to a lower-dimensional
non-linear manifold, while being more invariant to the vast majority of
directions orthogonal to the manifold. Finally, we show that by using the
learned features to initialize a MLP, we achieve state of the art
classification error on a range of datasets, surpassing other methods of
pre-training.

\end{abstract}

\section{Introduction}

A recent topic of interest\footnote{see NIPS'2010 Workshop on Deep Learning and
  Unsupervised Feature Learning} in the machine learning community is the
development of algorithms for unsupervised learning of a useful representation. 
This automatic discovery and extraction of features is often used in 
building a deep hierarchy of features, within the contexts of supervised,
semi-supervised, or unsupervised modeling.  See~\citet{Bengio-2009} for a
recent review of Deep Learning algorithms. Most of these methods exploit as basic
building block algorithms for learning one level of feature extraction: the
representation learned at one level is used as input for learning the
next level, etc.  The objective is that these representations become {\em better}
as depth is increased, but {\em what defines a good representation?} It is
fairly well understood what PCA or ICA do, but much remains to be done to understand
the characteristics and theoretical advantages of the representations learned by a Restricted
Boltzmann Machine~\cite{Hinton06}, an auto-encoder~\cite{Bengio-nips-2006},
sparse
coding~\cite{Olshausen-97,ranzato-07-small,Koray-08,Zeiler-cvpr2010}, or
semi-supervised embedding~\cite{WestonJ2008}. All of these produce a
non-linear representation which, unlike that of PCA or ICA,
can be stacked (composed) to yield deeper
levels of representation. It has also been observed
empirically~\cite{HonglakL2009} that the deeper levels often capture more
abstract features (such as parts of objects) defined in terms of less
abstract ones (such as sub-parts of objects or low-level visual features
like edges), and that these features are generally more
invariant~\cite{Goodfellow2009} to changes in the known factors of
variation in the data (such as geometric transformations in the case of
images). A simple approach, used here, to empirically verify that the learned
representations are useful, is to use them to initialize a classifier (such
as a multi-layer neural network), and measure classification error. Many
experiments show that deeper models can thus yield lower classification
error~\citep{Bengio-nips-2006,Jarrett-ICCV2009,VincentPLarochelleH2008-small}. 
%This
%simple and objective approach is also what we do here, although it would be
%nice to measure richer characteristics of the learned representations.

\myparagraph{Contribution.}
What principles should guide the learning of such intermediate representations?
They should capture as much as possible of the information in each given
example, when that example is likely under the underlying generating 
distribution. That is what auto-encoders~\cite{VincentPLarochelleH2008-small}
and sparse coding aim to achieve when minimizing reconstruction error.

We would also like these representations to be useful in characterizing
the input distribution, and that is what is achieved by
directly optimizing a generative model's likelihood (such as RBMs),
or a proxy, such as Score Matching~\cite{Hyvarinen-2005}.
In this paper, we introduce a penalty term that could be added
to either of the above contexts, which encourages the intermediate
representation to be robust to small changes of the input {\em
around the training examples}. We show through comparative experiments
on many benchmark datasets that this characteristic is useful to
learn representations that help training better classifiers.
Previous work has shown that deep learners can discover representations
whose features are invariant to some of the factors of variation
in the input~\citep{Goodfellow2009}. It would be nice to move
further in this direction, towards representation learning algorithms
which help to disentangle the factors of variation that underlie
the data generating distribution.
We hypothesize that whereas the proposed penalty term encourages
the learned features to be locally invariant without any preference
for particular directions, when it is combined with a reconstruction
error or likelihood criterion we obtain invariance in the directions
that make sense in the context of the given training data,
i.e., the variations that
are present in the data should also be captured in the learned
representation, but the other directions may be {\em contracted} in the learned
representation.

\section{How to extract robust features}
\label{sec:sensitivity}

% - explain how deep architectures extract robust features: Dumi's POV of regularization
%   and why it helps the generalization performs in a small paragraph
% - Explain why the denoising Auto-encoder achieves a robust feature extraction 
%   (manifold POV, Bishop's approximation)

% redondant avec ce qui précède
Most successful modern approaches for building deep networks begin by
initializing each layer in turn, using a local unsupervised learning
technique, to extract potentially useful features for the next layer.  When
used as feature extractors in this fashion, both RBMs and various flavors
of auto-encoders lead to a non-linear feature extractor of the exact same
form: a linear mapping followed by a sigmoid non-linearity\footnote{This
  corresponds to the encoder part of the traditional auto-encoder
  neural-network and its regularized variants. In RBMs, the conditional
  expectation of the hidden layer given the visible layer has the exact
  same form.}.  From this perspective, these algorithms are but different
unsupervised techniques to learn the parameters of a mapping of that form.
It is not yet fully understood what properties of such a mappings
contribute to superior classification performance (for classifiers
initialized with the produced features). It has been argued that mappings
that produce a sparse representation are to be encouraged, which inspired
several variants of sparse auto-encoders. 

The research we present here is motivated by a different property: our
working hypothesis is that a good representation of a likely input (under
the unknown data distribution) should be expected to remain rather stable
(i.e. be robust, invariant, insensitive) under tiny perturbations of that
input. This prompts us to propose an alternative regularization term for auto-encoders. 

To encourage robustness of the representation $f(x)$ obtained for a training input $x$ we
propose to penalize its {\em sensitivity} to that input, measured as the Frobenius norm
of the Jacobian $J_f(x)$ of the non-linear mapping. Formally, if input $x \in \R^{d_x}$ is
mapped by encoding function $f$ to hidden representation $h \in
\R^{d_h}$, this sensitivity penalization term is the sum of squares
of all partial derivatives of the extracted features with respect to input
dimensions:
\begin{equation}
\|J_f(x)\|_F^2 = \sum_{ij} \left( \frac{\partial h_j(x)}{\partial x_i} \right)^2.
\label{eq:sensitivity}
\end{equation}
Penalizing $\|J_f\|_F^2$ encourages the mapping to the feature space to be {\em
  contractive} in the neighborhood of the training data.  This geometric
perspective, which gives its name to our algorithm, will be further
elaborated on, in section~\ref{sec:geometry}, based on experimental evidence.
The {\em flatness} induced by having low valued first derivatives will
imply an {\em invariance} or {\em robustness} of the representation for
small variations of the input.
Thus in this study, terms like invariance, (in-)sensitivity, robustness,
flatness and contraction all point to the same notion.
 
While such a Jacobian term alone would encourage mapping to a useless
constant representation, it is counterbalanced in auto-encoder
training\footnote{Using also the now common additional constraint of
  encoder and decoder sharing the same (transposed) weights, which
  precludes a mere global contracting scaling in the encoder and expansion in the
  decoder.}  by the need for the learnt representation to allow a good
reconstruction of the input\footnote{A likelihood-related criterion would
also similarly prevent a collapse of the representation.}.

% JE PARLE DE CA PLUS BAS DANS LIENS ENTRE CAE ET AE+wd
% Note how
%in the linear case this criterion is just L2 penalization (weight decay),
%but the interesting aspects of it really arise in the non-linear case,
%where large weights can yield added robustness (as the non-linearity saturates).

%P: link with Tikhonov regularization (?)

%% With this perspective in mind, we will also experimentally compare, on
%% empirical data distributions, the expected sensitivity of the mappings
%% learnt by RBMs and by several forms of regularized auto-encoders that do
%% not explicitly optimize such a term.

%%%% P: DISCUSSION ANALYSE DE COURBE: barriere de grosse sensibilite entre le dataset et l'au-dela

% - explain that contraction should be used to regularize the  extraction
%   of features while performing reconstruction.

% - useful prior on the features (local invariance with respect to the input)

% - Distribution of the SV of the Jacobians
% - Average contraction using L1 and L2 norm for every pair (h(x),x)
% - Empirical measure to compare d(x1,x2) with d(h(x1),h(x2))
 
%\section{Traditional auto-encoders and a few variants}
\section{Auto-encoders variants}
\label{sec:autoencoders}

In its simplest form, an auto-encoder (AE) is composed of two parts, an encoder
and a decoder. It was introduced in the late 80's \cite{Rumelhart86b,Baldi89} as a technique
for dimensionality reduction, where the output of the encoder represents the 
reduced representation and where the decoder is tuned to reconstruct the initial input
from the encoder's representation through the minimization of a cost function.
More specifically when the encoding activation functions are linear
and the number of hidden units is inferior to the input dimension (hence forming
a bottleneck), it has been shown that the learnt parameters of the encoder 
are a subspace of the principal components of the input space \cite{Baldi89}.
With the use of non-linear activation functions an AE can however be expected to learn
more useful feature-detectors than what can be obtained with a simple PCA \citep{Japkowicz2000}. 
Moreover, contrary to their classical use as dimensionality-reduction techniques, in
their modern instantiation auto-encoders are often employed in a so-called {\em over-complete}
setting to extract a number of features larger than the input dimension,
yielding a rich higher-dimensional representation. In this setup, using some
form of regularization becomes essential to avoid uninteresting solutions
where the auto-encoder could perfectly reconstruct the input
without needing to extract any useful feature. 
This section formally defines the auto-encoder variants considered in this study.

%While traditional regularization approaches such as weight decay
%Beyond classical penalization of the squared norm of the auto-encoder's parameters 
%Obviously one needs to use some prior knowledge over the distribution
%of the weights, the input space, or on the joint of both to achieve a useful hidden
%representation.

\myparagraph{Basic auto-encoder (AE).}
The encoder is a function $f$ that maps an input  $x \in \R^{d_x}$
to hidden representation $h(x) \in \R^{d_h}$. It has the form
\begin{equation}
\label{eq:encoder}
h = f(x) = s_f(W x+b_h),
\end{equation}
where $s_f$ is a nonlinear {\em activation function}, typically a logistic $\mathrm{sigmoid}(z)=\frac{1}{1+e^{-z}}$. 
%P: TODO do we give results with other activation funcitons (rectifier, tanh, softplus)?
The encoder is parametrized by a $d_h \times d_x$ weight matrix $W$, and a bias vector $b_h \in \R^{d_h}$.

The decoder function $g$ maps hidden representation $h$ back to a reconstruction $y$:
\begin{equation}
\label{eq:encoder}
y = g(h) = s_g(W'h+b_y),
\end{equation}
where $s_g$ is the decoder's activation function, typically either the identity (yielding linear reconstruction) or a sigmoid. 
The decoder's parameters are a bias vector $b_y \in \R^{d_x}$, and matrix $W'$. 
In this paper we only explore the tied weights case, in which $W'=W^T$.

Auto-encoder training consists in finding parameters $\theta=\{W,b_h,b_y\}$
that minimize the reconstruction error on a training set of
examples $D_n$, which corresponds to minimizing the following objective function:

\begin{equation}
\label{eq:AE-objective}
\J_{\mathrm{AE}}(\theta) = \sum_{x \in D_n} L(x, g(f(x))),
\end{equation}

where $L$ is the reconstruction error. Typical choices include the squared error $L(x,y)=\|x-y\|^2$ used in cases of linear reconstruction 
and the cross-entropy loss when $s_g$ is the sigmoid (and inputs are in $[0, 1]$): $L(x,y)= - \sum_{i=1}^{d_x} x_i \log(y_i) + (1-x_i) \log (1-y_i)$.

\myparagraph{Regularized auto-encoders (AE+wd).}
The simplest form of regularization is {\em weight-decay} which favors small weights 
by optimizing instead the following regularized objective:
\begin{equation}
\label{eq:AEwd-objective}
\J_{\mathrm{AE+wd}}(\theta) = \left( \sum_{x \in D_n} L(x, g(f(x))) \right) + \lambda \sum_{ij} W_{ij}^2,
\end{equation}
where the $\lambda$ hyper-parameter controls the strength of the regularization.

% \myparagraph{Encouraging sparse representation:  (Sparse-AE)}

Note that rather than having a prior on what the weights should be, it is possible to
have a prior on what the hidden unit activations should be. From this
viewpoint, several techniques have been developed to encourage sparsity of
the representation ~\citep{koray-psd-08,HonglakL2008}. 

\iffalse
The
simplest approach, which we will use here to stay closest to the basic
auto-encoder training procedure, is to penalize the $L_1$ norm of the
representation, yielding the following regularized objective:

\begin{equation}
\label{eq:SparseAE-objective}
\J_{\mathrm{SparseAE}}(\theta) = \sum_{x \in D_n} L(x, g(f(x))) + \lambda \|f(x)\|_1 
\end{equation}
% Alternative form 
% - Sparse features Auto-encoders (L1 on activations) \\\\
% - Standard Weight decay AE\cite{?} \\\\
\fi

\myparagraph{Denoising Auto-encoders (DAE).}
A successful alternative form of regularization is obtained through the technique of
denoising auto-encoders (DAE) put forward by ~\citet{VincentPLarochelleH2008-small,Vincent-JMLR-2010}, where one simply
corrupts input $x$ before sending it through the auto-encoder, that is trained 
to reconstruct the clean version (i.e. to denoise). This yields the following objective function:

\begin{equation}
\label{eq:DAE-objective}
\J_{\mathrm{DAE}}(\theta) = \sum_{x \in D_n} \E_{\tilde{x}\sim q(\tilde{x}|x)} [ L(x, g(f(\tilde{x}))) ],
\end{equation}
where the expectation is over corrupted versions $\tilde{x}$ of examples $x$ obtained from a corruption process $q(\tilde{x}|x)$.
This objective is optimized by stochastic gradient descent (sampling corrupted examples).

Typically, we consider corruptions such as additive isotropic Gaussian noise:
$\tilde x = x+\epsilon, \epsilon \sim \mathcal{N}(0,\sigma^2I)$ and
a binary masking noise, where a fraction $\nu$ of input components
(randomly chosen) have their value set to 0.
The degree of the corruption ($\sigma$ or $\nu$) controls the degree of regularization.

% - small paragraph to motivate DAE citing Vincent et al(2008)
% - define their formalism and the different kind of noise distribution that can be used

\section{Contracting auto-encoders (CAE)}

From the motivation of robustness to small perturbations around the
training points, as discussed in section \ref{sec:sensitivity}, we propose
an alternative regularization that favors mappings that are more 
strongly contracting at the training samples (see section \ref{sec:geometry} for a longer
discussion).
The Contracting auto-encoder (CAE) is obtained with the regularization term of eq.~\ref{eq:sensitivity} yielding objective function

\begin{equation}
\label{eq:CAE-objective}
\J_{\mathrm{CAE}}(\theta) = \sum_{x \in D_n} \left( L(x, g(f(x))) + \lambda \|J_f(x)\|_F^2 \right) 
\end{equation}

\myparagraph{Relationship with weight decay.}
It is easy to see that the squared Frobenius norm of the Jacobian corresponds to a
L2 weight decay in the case of a {\em linear encoder} (i.e. when $s_f$ is the
identity function).  In this special case $\J_{\mathrm{CAE}}$ and
$\J_{\mathrm{AE+wd}}$ are identical.  Note that in the linear case, keeping
weights small is the only way to have a contraction. But with a sigmoid
non-linearity, contraction and robustness can also be achieved by
driving the hidden units to their saturated regime.

\myparagraph{Relationship with sparse auto-encoders.}
Auto-encoder variants that encourage sparse representations aim at having,
for each example, a majority of the components of the representation close
to zero. For these features to be close to zero, they must have been
computed in the left saturated part of the sigmoid nonlinearity, which 
is almost flat, with a tiny first derivative. This yields a corresponding
small entry in the Jacobian $J_f(x)$.  Thus, sparse auto-encoders that output
many close-to-zero features, are likely to correspond to a highly
contractive mapping, even though contraction or robustness are not {\em
  explicitly} encouraged through their learning criterion.

% P: TODO in experimental section, make sure we say best sparse-AE did not select high degrees of sparsity (louche, peut-être que de meilleurs algos l'auraient fait)

\myparagraph{Relationship with denoising auto-encoders.}
Robustness to input perturbations was also one of the motivation of the
denoising auto-encoder, as stated in \citet{Vincent-JMLR-2010}. 
The CAE and the DAE differ however in the following ways:
\begin{itemize}
\item CAEs explicitly encourage robustness of representation $f(x)$,
  whereas DAEs encourages robustness of reconstruction $(g \circ f)(x)$
  (which may only partially and indirectly encourage robustness of the
  representation, as the invariance requirement is shared between the two parts of the 
  auto-encoder).  We believe that this property make CAEs a better choice
  than DAEs to learn useful feature extractors. Since we will use
  only the encoder part for classification, robustness of the extracted 
  features appears more important than robustness of the reconstruction.
\item DAEs' robustness is obtained {\em stochastically} (eq.~\ref{eq:DAE-objective}) by having several explicitly
  corrupted versions of a training point aim for an identical reconstruction.
  By contrast, CAEs' robustness to tiny perturbations is obtained {\em analytically} by penalizing
  the magnitude of first derivatives $\|J_f(x)\|_F^2$ at training points only (eq.~\ref{eq:CAE-objective}).
\end{itemize}

\subsection{Analytical link between denoising auto-encoders and contracting auto-encoders}

If the noise used in the DAE is Gaussian, its effect can be approximated analytically with an added penality term on a standard autoencoder 
cost function\citet{bishop95training} and \citet{Kingma+LeCun-2010}. Let us define $\mathcal{L}(\theta,x)$ as the loss function of the auto-encoder. We can write 
the expected cost of our loss as:
\begin{equation}
\label{eq:expcost}
\mathcal{C}(\theta) = \int \mathcal{L}(x,\theta)p(x)dx
\end{equation}

The empirical cost can be written by expressing our probabilty distribution as a series of dirac functions centered on our samples:

\begin{equation}
\label{eq:cleancost}
\mathcal{C}^{\mathrm{clean}}(\theta) = \int \mathcal{L}(x,\theta)\delta(x_i-x)dx = \frac{1}{n}\sum_i^n \mathcal{L}(x_i,\theta)
\end{equation}

If we were to express out density $p(x)$ as a series of Gaussian kernels centered on our samples and with diagonal covariance matrix, our empirical cost function would become:

\begin{equation}
\label{eq:noisycost}
\mathcal{C}^{\mathrm{noisy}}(\theta) = \frac{1}{n}\sum_i^n \int \mathcal{L}(x,\theta)\mathcal{N}_{x_i,\sigma^2}(x)dx
\end{equation}

In the context of a de-noising auto-encoder, we approximate this cost by using samples corrupted with a Gaussian noise. 
We can express the difference between these two costs as a function. Our goal is to find an approximation for this function.

\begin{equation}
\label{eq:penalty}
\mathcal{C}^{\mathrm{noisy}}(\theta) = \mathcal{C}^{\mathrm{clean}}(\theta) + \phi(\theta)
\end{equation}

and thus:

\begin{equation}
\label{eq:difference}
\phi(\theta)= \frac{1}{n}\sum_i^n \left[ \int \mathcal{L}(x,\theta)\mathcal{N}_{x_i,\sigma^2}(x)dx - \mathcal{L}(x_i,\theta) \right]
\end{equation}

We can write the term inside the integral by defining the noise term $\varepsilon = x - x_i$:

\begin{equation}
\label{eq:simplediff}
D =  \int \mathcal{L}(x_i + \varepsilon)\mathcal{N}_{0,\sigma^2}(\varepsilon)d\varepsilon  -\mathcal{L}(x_i)
\end{equation}

We can approximate the first term by a Taylor series:
\begin{equation}
\label{eq:taylor}
\mathcal{L}(x + \varepsilon) =  \mathcal{L}(x)  + \langle J_{\mathcal{L}}(x), \varepsilon \rangle +\frac{1}{2} \varepsilon^T . H_{\mathcal{L}}(x) . \varepsilon + o(\varepsilon)\
\end{equation}

By using this approximation and simplifying our integral, we obtain the following expression for our function:

\begin{equation}
\label{eq:phi}
\phi(\theta) \approx \frac{\sigma^2}{2n}\sum_i^n \trace(H_{\mathcal{L}}(x_i))
\end{equation}

\begin{equation}
\label{eq:final_eq}
\mathcal{C}^{\mathrm{noise}}(\theta) \approx  \mathcal{C}^{\mathrm{clean}}(\theta) + \frac{\sigma^2}{2n}\sum_i^n \trace(H_{\mathcal{\bar{C}}}(\theta))
\end{equation}

The de-noising autoencoder cost function can thus be approximated by using a classical auto-encoder with a penality on the trace of the Hessian of the cost fucntion
versus the inputs. When the cost function is the MSE of the reconstruction, we can write the Hessian as:

\begin{equation}
\begin{split}
 H_{L}(x) & = \frac{\partial}{\partial x} \left(\frac{\partial}{\partial x} \left( \parallel g(f(x)) - x \parallel^2 \right)\right) \\
 & = 2   \frac{\partial}{\partial x} \left( \frac{\partial g\left(f(x)\right)}{\partial x}  \left(  g(f(x)) - x \right)\right) \\
 & = 2  H_{g\circ f}(x)   \left(  g(f(x)) - x \right) + 2\left<J_{g\circ f}(x)^T, J_{g\circ f}(x)\right> \\
\end{split}
\end{equation}

By taking the trace of the above results, we get:

\begin{equation}
\trace \left( H_{L}(x) \right) = 2  \left(g(f(x)) - x\right) \trace\left(H_{g\circ f}(x)\right)   + 2\left|\left|  J_{g\circ f}(x) \right|\right|_F^2
\end{equation}

%SO WHAT? CONCLUDE WHAT DO WE LEARN FROM THIS?
The first term of the equation scales with the reconstruction cost and will diminish accordingly. The second term of the
equation is the Froebenius norm of the Jacobian. Note however that the Jacobian in this case is on the $reconstruction$ and not on the $representation$
as with our proposed penality.

\myparagraph{Why $J_{f}(x)$ is a better choice than $J_{g\circ f}(x)$ .}
% Two main reasons:
%  - Shared invariance between encoder/decoder
%  - Saturation leads to hard optimization (local minima, blocks gradients)
Adding Gaussian noise during the training of an auto-encoder is thus asymptotically equivalent to adding a 
regularization term to the objective function. In the DAE setting with MSE cost, the penalty term is the norm of the Jacobian of the 
reconstruction units with respect to the input units. This encourages the output to be invariant to small changes in the inputs. Note that
{\em the invariance requirement is shared between the two parts of the 
auto-encoder} and not explicitly on the $representation$. If the goal of the auto-encoder is to initialise the weights of a Multi-Layer Perceptron (MLP),
we do not care about the invariance of the $reconstruction$ since we only use the $representation$. We would like the invariances captured by the autoecnoder to be predominantly on the
$representation$. By using as a penality the norm of the Jacobian of the $representation$, we are doing this explicitly.

The other drawback of having a regularization over $J_{g\circ f}(x)$ comes from an
optimization point of view. The gradient of the total cost function with respect to the parameters
of the encoder $W$ is a function of $J_{g}(f)$.  Since 
$J_{g\circ f}(x) = J_{g}(f) . J_{f}(x)$, minimizing $J_{g\circ f}(x)$ by reducing $J_{g}(f)$ could lead to difficulties
in minimizing the reconstruction cost.

%Another point is that the most commonly used activation functions
%in neural networks are saturating functions, and the penalty term is in part satisfied when driving
%the activations to their saturated regime. For a binary dataset this could be harmless, but
%when the inputs are dense in some interval, this could lead to an instability in the reconstruction process
%and degrade the quality of the extracted features

% \myparagraph{Locally invariant features} (is it necessary?)

%We show that feature extraction is mostly beneficial when the learnt features are
%invariant to a local transformation of the input. Vincent et al.\cite{?} showed that
%DAE yield robust feature extraction that outperforms all other forms of AE,
%but did not really elaborate the idea of regularization induced by the added noise.
%Using Bishop's approximation \cite{bishop95training} of the noisy objective function, we derived the added
%penalty term induced by the DAE, helping us to understand what is explicitly happening
%during the optimization process and the effects of noise over the learnt parameters of
%the AE.

% \subsection{Advantages of the CAE}

% \myparagraph{Toward binary activations.}

% - Discuss the penalty term and it's effect on the parameters when using a
%   sigmoidal non-linearity
% - Explanatory graph to explain

% \myparagraph{Sparsity.} 

% - Xavier G. is going to verify this hypothesis

% - Ask Xavier G. to find an explanation

\section{Experiments and results}

\myparagraph{Considered models.}
In our experiments, we compare the proposed Contracting Auto Encoder (CAE) against the following models for 
unsupervised feature extraction:

\begin{myitemize}
\item RBM-binary : Restricted Boltzmann Machine trained by Contrastive Divergence, 
\item AE: Basic auto-encoder,
\item AE+wd: Auto-encoder with weight-decay regularization,
\item DAE-g: Denoising auto-encoder with Gaussian noise,
\item DAE-b: Denoising auto-encoder with binary masking noise,
%\item CAE: Contracting auto-encoder.
\end{myitemize}

All auto-encoder variants used tied weights, a sigmoid activation function
for both encoder and decoder, and a cross-entropy reconstruction error (see
Section~\ref{sec:autoencoders}). They were trained by optimizing their (regularized) 
objective function on the training set by stochastic gradient descent. 
As for RBMs, they were trained by Contrastive Divergence.

These algorithms were applied on the training set without using the labels
(unsupervised) to extract a first layer of features. Optionally the
procedure was repeated to stack additional feature-extraction layers on top of
the first one.
Once thus trained, the learnt parameter values of the resulting
feature-extractors (weight and bias of the encoder) were used as
initialisation of a multilayer perceptron (MLP) with
an extra random-initialised output layer. The whole network was then
{\em fine-tuned} by a gradient descent on a supervised objective appropriate for
classification~\footnote{We used sigmoid+cross-entropy for binary
  classification, and log of softmax for multi-class problems}, using the
labels in the training set.

\myparagraph{Datasets used.}
We have tested our approach on a benchmark of image classification problems, namely:
\begin{myitemize}
\item{\em CIFAR-10}: the image-classification task ($32\times 32 \times 3$ channels RGB) \citep{KrizhevskyHinton2009}. 
\item{\em CIFAR-bw}: a gray-scale version of the original CIFAR-10. The gray-scale versions were obtained with a color weighting of 0.3 for red, 0.59 for green and
0.11 for blue.
\item \mnistfull : the well-known digit classification problem ($28\times 28$ gray-scale pixel values scaled to [0,1]). It has 50000 examples for training, 10000 for validation, and 10000 for test.
%\item{\em CIFAR-bw}: a grayscale version of the CIFAR-10 \cite{KrizhevskyHinton2009} image-classification task. The grayscale versions were obtained with a color weighting of 0.3 for red, 0.59 for green and
%0.11 for blue.
%\item \mnistfull : the well-known~\cite{LeCun+98} digit classification problem ($28\times 28$ grayscale pixel values scaled to [0,1]). It has 50000 examples for training, 10000 for validation, and 10000 for test.
\end{myitemize}

Six harder digit recognition problems used in the benchmark
of~\citet{larochelle-icml-2007}\footnote{Datasets available at \url{http://www.iro.umontreal.ca/~lisa/icml2007}.}. 
They were
derived by adding extra factors of variation to MNIST digits.  Each has
10000 examples for training, 2000 for validation, 50000 for test.
\begin{myitemize}
\item \mnistbasic : Smaller subset of MNIST. 
\item \mnistrot : digits with added random rotation. 
\item \mnistbackrand : digits with random noise background. 
\item \mnistbackimage : digits with random image background.
\item \mnistrotbackimage : digits with rotation and image background.
\end{myitemize}

Two artificial shape classification problems from the benchmark of~\citet{larochelle-icml-2007}:
\begin{myitemize}
\item \rectangles : Discriminate between tall and wide rectangles (white on black).
\item \rectanglesimage : Discriminate between tall and wide rectangular image on a different background image.
\end{myitemize}

\subsection{Classification performance}

\subsubsection{MNIST and CIFAR-bw}
Our first series of experiments focuses on the MNIST and CIFAR-bw
datasets. We compare the classification performance obtained by a neural
network with one hidden layer of 1000 units, initialized 
with each of the unsupervised algorithms under consideration.  For each case, 
we selected the value of hyperparameters (such as the strength of
regularization) that yielded, after supervised fine-tuning, the best
classification performance on the validation set. 
Final classification error rate was then computed on the test set.
With the parameters obtained after unsupervised pre-training (before fine-tuning), 
we also computed in each case the average value of the encoder's contraction $\|J_f(x)\|_F$ 
on the validation set, as well as a measure of the average fraction of
saturated units per example\footnote{We consider a unit saturated if its
  activation is below 0.05 or above 0.95. Note that in general the set of saturated units is expected to vary with each example.}. 
These results are reported in Table~\ref{tab:bench-results-mnist-cifar}.
We see that the local contraction measure (the average $\|J_f\|_F$) on the
pre-trained model {\bf strongly correlates} with the final classification
error. The CAE, which explicitly tries to minimize this measure while
maintaining a good reconstruction, is the best-performing model.
datasets.

%\footnote{e.g. the optimal
%  unsupervised learning rate was found to be in range [0.0005,0.02]
%  depending on the dataset and model.}

% - classification on mnist (DAE-1 bin, CAE-1, MLP-1, Regularized AE) vs contraction measure
\begin{table}
\begin{center}
\begin{tabular}{c|l|r|r|r|} % \cline{2-5}
& \multirow{2}{*}{\bf Model}       &  {\bf Test}    & {\bf Average}                   & \multirow{2}{*}{\bf SAT}           \\ 
&                   &  {\bf error}         & {\bf $\|J_f(x)\|_F$}            &           \\\hline \hline
\multirow{6}{*}{\rotatebox{90}{\bf MNIST}}
& CAE               &{\bf 1.14}            & 0.73 {\scriptsize $10^{-4}$}    & 86.36\%              \\\cline{2-5}
& DAE-g             &{\bf 1.18}            & 0.86 {\scriptsize $10^{-4}$}    & 17.77\%              \\\cline{2-5}
& RBM-binary        &1.30                  & 2.50 {\scriptsize $10^{-4}$}    & 78.59\%             \\\cline{2-5}
& DAE-b             &1.57                  & 7.87 {\scriptsize $10^{-4}$}    & 68.19\%                    \\\cline{2-5}
& AE+wd             &1.68                  & 5.00 {\scriptsize $10^{-4}$}    & 12.97\%              \\\cline{2-5}
& AE                &1.78                  & 17.5 {\scriptsize $10^{-4}$}    & 49.90\%             \\ \hline \hline
\multirow{5}{*}{\rotatebox{90}{\bf CIFAR-bw}}
& CAE               &{\bf 47.86}          & 2.40 {\scriptsize $10^{-5}$}    & 85,65\%   \\  \cline{2-5} 
& DAE-b             &{\bf 49.03}          & 4.85 {\scriptsize $10^{-5}$}    & 80,66\%   \\  \cline{2-5}
& DAE-g             &54.81                & 4.94 {\scriptsize $10^{-5}$}    & 19,90\%   \\  \cline{2-5}
& AE+wd             &55.03                & 34.9 {\scriptsize $10^{-5}$}    & 23,04\%   \\  \cline{2-5}
& AE                &55.47                & 44.9 {\scriptsize $10^{-5}$}    & 22,57\%   \\  \hline
\end{tabular}
\end{center}
\caption{Performance comparison of the considered models on MNIST (top half) and CIFAR-bw (bottom half). 
Results are sorted in ascending order of classification error on the test set.
Best performer and models whose difference with the best performer was not statistically significant are in bold.
Notice how the average Jacobian norm (before fine-tuning) appears correlated with the final test error. 
SAT is the average fraction of saturated units per example. 
Not surprisingly, the CAE yields a higher proportion of saturated units. }
\label{tab:bench-results-mnist-cifar}
\end{table}

%\subsection{CAEs for pretraining deep networks}

%% In our next series of experiments, we followed a similar procedure to
%% \citet{larochelle-icml-2007,VincentPLarochelleH2008-small,Vincent-JMLR-2010},
%% stacking two layers of CAEs to initialize a MLP that is then fine-tuned.
%% Performance of two-layer stacked CAEs is compared to that of other deep
%% models on the benchmark datasets
%% of~\citet{larochelle-icml-2007,Vincent-JMLR-2010}.  
Results given in Table~\ref{tab:bench-results} compare the performance of stacked CAEs 
on the benchmark problems of \citet{larochelle-icml-2007} to the three-layer models 
reported in \citet{Vincent-JMLR-2010}. Stacking a second layer CAE on top
of a first layer appears to significantly improves performance, thus demonstrating
their usefulness for building deep networks. Moreover on the majority of
datasets, 2-layer CAE beat the state-of-the-art 3-layer model.

\subsubsection{CIFAR-10}

The pipeline of preprocessing steps we used here is similar to \cite{?}.
We randomly extracted $160 000$ patches $8\times 8$ from the $10 000$ first images of CIFAR-10.
For each patch, we substract the mean and divide by the standard deviation (
local contrast normalization). Then, a PCA is fitted on this set of patches. The 2 first
components (corresponding to black patches) are dropped but we kept the next 80 first components
(over 192). For building the final training set of patches, we project these patches on the PCA components,
perform whitening i.e divide by the eigen values, and pass it through a logistic funtion
in order to map it to $[0,1]$.

A Contracting-Auto-Encoder with a number of hidden units $n_{hid}\in \lbrace 50, 100, 200, 400 \rbrace$ 
is trained on this set by minimizing the cross-entropy reconstruction error and the regularizer with
stochastic gradient descent. We present some filters learned during this process in Figure XX.

Finally, we evaluate the classification performance of our algorithm with a linear classifier.
The preprocessing steps are applied convolutionnaly with a stride equal to one to get $n_{hid}$ feature
maps of size $25\times 25$. Features are sum-pooled together over the quadrants of the feature maps.
By applying this coarse dimensionality reduction technique, we obtain features of dimension 
$4 n_{hid}$. We fed a linear $L2$-regularized SVM with these features and reported the test classification
accuracy in TAB XX. $L2$ regularizer was chosen using a 5-fold cross validation.

\subsection{Closer examination of the contraction}

\begin{figure}[h]
\vskip 0.2in
\begin{center}
\includegraphics[width=\columnwidth]{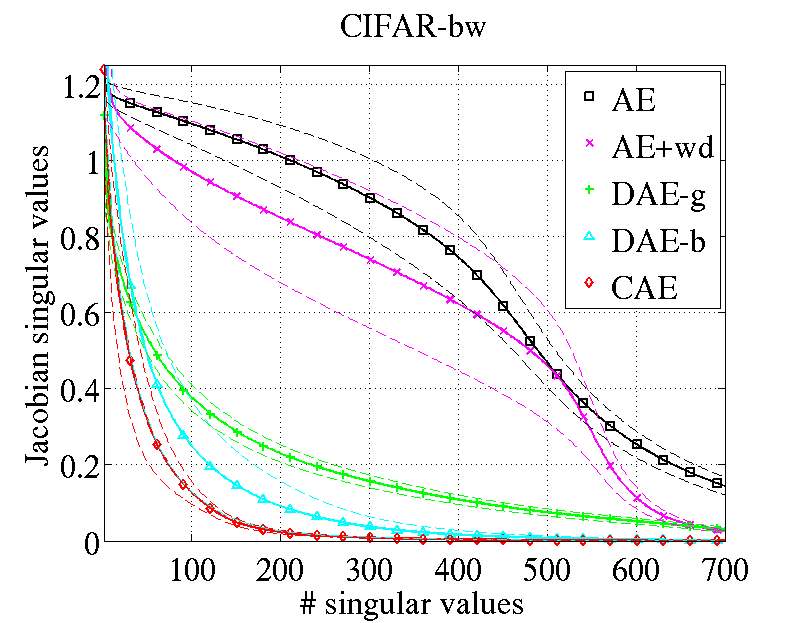}
\caption{Average spectrum of the encoder's Jacobian, for the CIFAR-bw dataset. Large
singular values correspond to the local directions of ``allowed'' variation learnt from the dataset.
The CAE having fewer large singular values and a sharper decreasing spectrum, 
it suggests that it does a better job of characterizing a {\em low-dimensional manifold} 
near the training examples.}
\label{fig:cifarbw_svd}
\end{center}
\vskip -0.2in
\end{figure}

To better understand the feature extractor produced by each
algorithm, in terms of their contractive properties, we used the following analytical tools: 

\myparagraph{What happens locally: looking at the singular values of the Jacobian.}
A high dimensional Jacobian contains directional information: the amount of
contraction is generally not the same in all directions. This can be
examined by performing a singular value decomposition of $J_f$.  We
computed the average singular value spectrum of the Jacobian over the
validation set for the above models. 
Results are shown in Figure~\ref{fig:cifarbw_svd} and will be 
discussed in section~\ref{sec:geometry}.

\myparagraph{What happens further away: contraction curves.}
The Frobenius norm of the Jacobian at some point $x$ measures the contraction of the
mapping {\em locally} at that point.
Intuitively the contraction induced by the proposed penalty term
can be measured beyond the immediate training examples,  
by the ratio of the distances between two  
points in their original (input) space and their distance once mapped in the feature space.
We call this measure {\em contraction ratio}.
In the limit where the variation in the input space is infinitesimal,
this corresponds to the derivative (i.e. Jacobian) of the representation map.

\begin{figure}[h]
\vskip 0.2in
\begin{center}
\includegraphics[width=\columnwidth]{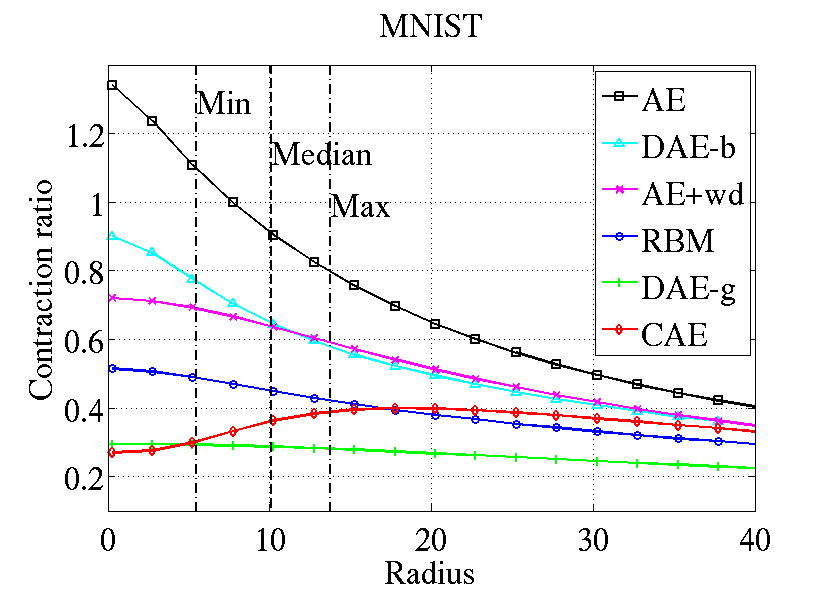}
\includegraphics[width=\columnwidth]{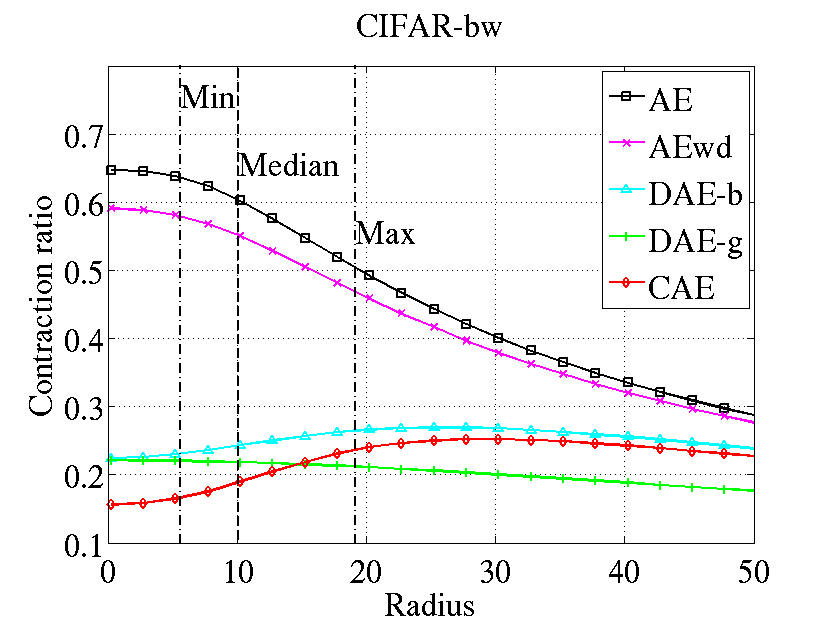}
\caption{Contraction curves obtained with the considered models on MNIST (top) and CIFAR-bw (bottom). 
See the main text for a detailed interpretation.}
\label{fig:l2_contraction}
\end{center}
\vskip -0.2in
\end{figure}

For any encoding function $f$, we can measure the {\em average contraction
  ratio} for pairs of points, one of which, $x_0$ is picked from the
validation set, and the other $x_1$ randomly generated on a sphere of
radius $r$ centered on $x_0$ in input space. How this average ratio evolves
as a function of $r$ yields a {\em contraction curve}. We have computed
these curves for the models for which we reported classification
performance (the contraction curves are however computed with their initial
parameters prior to fine tuning). Results are shown in Figure~\ref{fig:l2_contraction}
for single-layer mappings and in Figure~\ref{fig:l2_contraction_layer} for 2 and 3 layer mappings.
They will be discussed in detail in the next section.

% - exp measure on mnist for all XG models (graph)  (XM)
% - exp measure on cifarbw for all XG models (graph) (XM)
% - exp measure on icml CAE-1 vs MLP only 1-layer vs CAE-2 vs CAE-3 
%   last layer( time is a constraint, can't do more models) (graph) (XM)

% - svd measure on mnist for all XG models (graph)  (XG)
% - svd measure on cifarbw for all XG models (graph) (XG)
% - svd measure on icml CAE vs MLP only 1-layer vs vs CAE-2 CAE-3 
%   last layer( time is a constraint, can't do more models) (graph) (XG)

% - jacobi avg L1 measure on mnist for all XG models (graph)  (SR)
% - jacobi avg L1 measure on cifarbw for all XG models (graph) (SR)
% - jacobi avg L1 norm measure on icml CAE vs MLP only 1-layer vs
%   CAE-2 CAE-3 last layer( time is a constraint, can't do more models) (graph) (SR)

\begin{table*}[t!]
\begin{center}
\resizebox{12cm}{!} {
\begin{tabular}{|l|r|r|r|r|r|r|r|r|} \hline
{\bf Data Set}      & $\mathbf{SVM}_{rbf}$    & {\bf SAE-3}       & {\bf RBM-3}            & {\bf DAE-b-3}          & {\bf CAE-1}            & {\bf CAE-2}  \\\hline \hline
\mnistbasic        & {\bf 3.03}\mypm{0.15}   & 3.46\mypm{0.16}   & 3.11\mypm{0.15}        &  2.84\mypm{0.15}       &  2.83\mypm{0.15}       & {\bf 2.48}\mypm{0.14} \\ \hline
\mnistrot          &  11.11\mypm{0.28}       & 10.30\mypm{0.27}  & 10.30\mypm{0.27}       & {\bf 9.53}\mypm{0.26}  &  11.59\mypm{0.28}      & {\bf 9.66}\mypm{0.26}\\\hline
\mnistbackrand     &  14.58\mypm{0.31}       & 11.28\mypm{0.28}  & {\bf 6.73}\mypm{0.22}  & 10.30\mypm{0.27}       &  13.57\mypm{0.30}      & 10.90 \mypm{0.27}  \\ \hline
\mnistbackimage    & 22.61\mypm{0.379}       & 23.00\mypm{0.37}  &  16.31\mypm{0.32}      & 16.68\mypm{0.33}       &  16.70\mypm{0.33}      & {\bf 15.50}\mypm{0.32}\\ \hline
\mnistrotbackimage &  55.18\mypm{0.44}       & 51.93\mypm{0.44}  & 47.39\mypm{0.44}       & {\bf 43.76}\mypm{0.43} &  48.10\mypm{0.44}      &  45.23\mypm{0.44}\\\hline
\rectangles        & {\bf 2.15}\mypm{0.13}   & 2.41\mypm{0.13}   & 2.60\mypm{0.14}        & 1.99\mypm{0.12}        & 1.48\mypm{0.10}        & {\bf 1.21}\mypm{0.10}  \\ \hline
\rectanglesimage   &  24.04\mypm{0.37}       &  24.05\mypm{0.37} & 22.50\mypm{0.37}       & {\bf 21.59}\mypm{0.36} & {\bf 21.86}\mypm{0.36} & {\bf 21.54}\mypm{0.36}\\ \hline

% results with original biased methodology (choose single best hyper-parm on k-fold valid)
% \tzanetakis        & {\bf 14.41}\mypm{2.18} & 18.15\mypm{1.43}        & {\bf 16.94}\mypm{1.95} & 18.21\mypm{0.85}       & {\bf 16.68}\mypm{1.30}(0.05) \\ \hline
\end{tabular}
}
\end{center}
\caption{Comparison of stacked contracting auto-encoders with 1 and 2
  layers (CAE-1 and CAE-2) with other 3-layer stacked models and baseline
  SVM. Test error rate on all considered classification problems is
  reported together with a 95\% confidence interval. Best performer is in
  bold, as well as those for which confidence intervals overlap.  Clearly
  CAEs can be successfully used to build top-performing deep networks. 2
  layers of CAE often outperformed 3 layers of other stacked models.  }
\label{tab:bench-results}
\end{table*}

\subsection{Discussion: Local Space Contraction}
\label{sec:geometry}

From a geometrical point of view, the robustness of the features can be seen
as a contraction of the input space when projected in the feature space,
in particular {\em in the neighborhood of the examples from the data-generating
distribution}: otherwise (if the contraction was the same at all distances)
it would not be useful, because it would just be a global scaling.
This is happening with the proposed penalty, but rarely so without it,
as illustrated on the contraction curves of Figure~\ref{fig:l2_contraction}.
For all algorithms tested except the proposed CAE and the Gaussian corruption DAE
(DAE-g), the contraction ratio {\em decreases} (i.e., towards more contraction) as we move away from the training 
examples (this is due to more saturation, and was expected), 
whereas for the CAE the contraction ratio {\em initially increases}, up to the point
where the effect of saturation takes over (the bump occurs at about the
maximum distance between two training examples).

Think about the case where the training examples congregate near a low-dimensional
manifold. The variations present in the data (e.g. translation and rotations
of objects in images) correspond to local dimensions along
the manifold, while the variations that are small or rare in the data correspond to the directions
orthogonal to the manifold (at a particular point near the manifold, corresponding
to a particular example). The proposed criterion is trying to make the
features invariant in all directions around the training examples, but
the reconstruction error (or likelihood) is making sure that that the
representation is faithful, i.e., can be used to reconstruct the input
example. Hence the directions that {\em resist} to this contracting pressure
(strong invariance to input changes) are the directions
present in the training set. Indeed, {\em if the variations along these
directions present in the training set were not preserved, neighboring
training examples could not be distinguished and properly reconstructed.} 
Hence the directions where the contraction is strong 
(small ratio, small singular values of the Jacobian matrix)
are also the directions where the model believes that the input density
drops quickly, whereas the directions where the contraction is weak (closer to 1,
larger contraction ratio, larger singular values of the Jacobian matrix)
correspond to the directions where the model believes that the input
density is flat (and large, since we are near a training example).

We believe that this contraction penalty thus helps 
the learner carve a kind of mountain
supported by the training examples, and generalizing to a ridge between them.
What we would like is for these ridges to correspond to some
directions of variation present in the data, associated to underlying
factors of variation. 
How far do these ridges extend around each training example
and how flat are they? This can be visualized comparatively 
with the analysis of Figure~\ref{fig:l2_contraction}, 
with the contraction ratio for different distances from the training examples.

Note that different features (elements of
the representation vector) would be expected to have ridges
(i.e. directions of invariance) in different directions, and that the
``dimensionality'' of these ridges (we are in
a fairly high-dimensional space) gives a hint as to the local dimensionality
of the manifold near which the data examples congregate.
The singular value spectrum of the Jacobian
informs us about that geometry. The number
of large singular values should reveal the dimensionality of
these ridges, i.e., of that manifold near which examples concentrate. 
This is illustrated in Figure~\ref{fig:cifarbw_svd}, showing
the singular values spectrum of the encoder's Jacobian. The CAE by far does
the best job at representing the data variations near a lower-dimensional
manifold, and the DAE is second best, while ordinary auto-encoders
(regularized or not) do not succeed at all in this respect.

\begin{figure}[h]
\vskip 0.2in
\begin{center}
\includegraphics[width=\columnwidth]{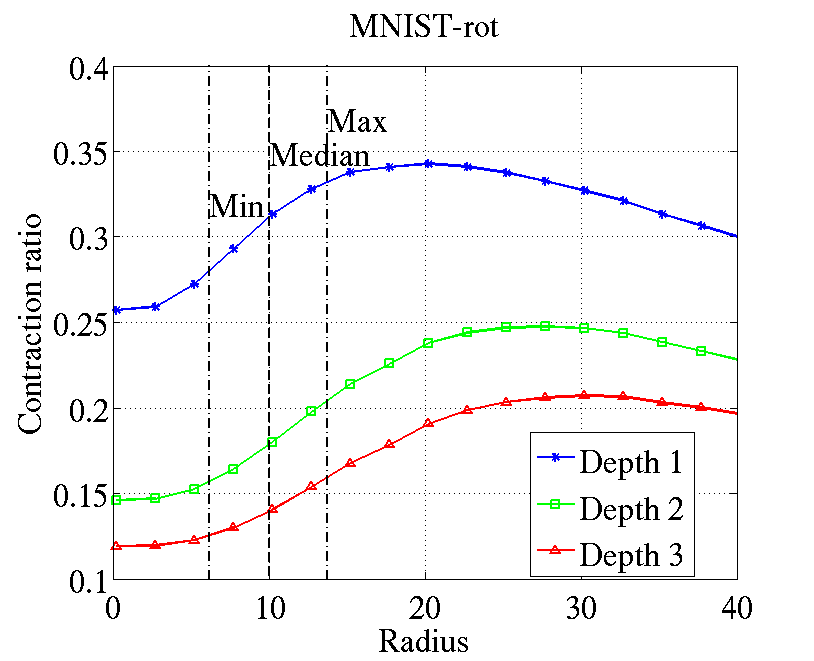}
\caption{Contraction effect as a function of depth. Deeper encoders produce features that
are more invariant, over a farther distance, corresponding to flatter ridge of the density
in the directions of variation captured.}
\label{fig:l2_contraction_layer}
\end{center}
\vskip -0.2in
\end{figure}

What happens when we stack a CAE on top of another one, to build
a deeper encoder? This is illustrated in Figure~\ref{fig:l2_contraction_layer},
which shows the average contraction ratio for different distances around
each training point, for depth 1 vs depth 2 encoders.Composing two CAEs yields even more contraction
and even more non-linearity, i.e. a sharper profile, with a flatter
level of contraction at short and medium distances, and a delayed
effect of saturation (the bump only comes up at farther distances).
We would thus expect higher-level features to be more invariant
in their feature-specific directions of invariance, which is exactly the kind
of property that motivates the use of deeper architectures.

% - classification on larochelle-icml-2007 (CAE-1, CAE-2, CAE-3 vs icml benchmarks)

\section{Conclusion}

In this paper, we attempt to answer the following question: {\em what makes a good
  representation?}.  Besides being useful for a particular task, which we
can measure, or towards which we can train a representation, this paper
highlights the advantages for representations to be {\em locally invariant
  in many directions} of change of the raw input. This idea is implemented
by a penalty on the Frobenius norm of the Jacobian matrix of the encoder
mapping, which computes the representation. The paper also introduces
empirical measures of robustness and invariance, based on the contraction
ratio of the learned mapping, at different distances and in different
directions around the training examples. We hypothesize that this reveals
the manifold structure learned by the model, and we find (by looking at the
singular value spectrum of the mapping) that the Contracting
Auto-Encoder discovers lower-dimensional manifolds. In addition,
experiments on many datasets suggest that this penalty always helps an
auto-encoder to perform better, and competes or improves upon the
representations learned by Denoising Auto-Encoders or RBMs, in terms of
classification error.

{%\small
\bibliography{strings,strings-shorter,ml,myrefs,aigaion-shorter}
\bibliographystyle{natbib}
}

\end{document}